\documentclass[11pt,a4paper,twocolumn]{article}
\usepackage[margin=22mm]{geometry}
\usepackage[T1]{fontenc}
\usepackage{lmodern}
\usepackage{microtype}
\usepackage{amsmath}
\usepackage{booktabs}
\usepackage[table]{xcolor}
\usepackage{graphicx}
\usepackage[font=small,labelfont=bf,aboveskip=5pt,belowskip=0pt]{caption}
\usepackage{stfloats}
\usepackage{flushend}
\usepackage{enumitem}
\usepackage[numbers,sort&compress]{natbib}
\usepackage[hidelinks]{hyperref}

\title{Agentic Waterfall: An Intent-Driven Information Discovery Framework for Reclaiming User Control in the Attention Economy}
\author{Yige Luo, Shining Zhang, Jikun Liu, Hongjiao Qiang, Ran Guan\\[2pt]\normalsize 2012 Laboratories, Huawei}
\date{\small November 2026}
\hypersetup{pdftitle={Agentic Waterfall: An Intent-Driven Information Discovery Framework for Reclaiming User Control in the Attention Economy},pdfauthor={Yige Luo, Shining Zhang, Jikun Liu, Hongjiao Qiang, Ran Guan}}
\definecolor{waterfallrow}{HTML}{EBF3FB}
\definecolor{urlblue}{HTML}{0055AA}

\renewenvironment{abstract}{%
  \par\vspace{6pt}%
  \centerline{\large\bfseries Abstract}%
  \vspace{6pt}%
  \begin{list}{}{%
    \setlength{\leftmargin}{1.1em}%
    \setlength{\rightmargin}{1.1em}%
    \setlength{\topsep}{0pt}%
    \setlength{\partopsep}{0pt}%
    \setlength{\parsep}{0pt}%
  }\item\relax\small\linespread{1.055}\selectfont
}{%
  \end{list}\par\vspace{10pt}%
}

\begin{document}
\flushbottom
\renewcommand{\dbltopfraction}{0.9}
\renewcommand{\textfraction}{0.08}
\maketitle
\begingroup
\setlength{\textfloatsep}{6pt plus 2pt minus 2pt}
\setlength{\dbltextfloatsep}{6pt plus 2pt minus 2pt}
\setlength{\dblfloatsep}{6pt plus 2pt minus 2pt}
\setlength{\floatsep}{8pt plus 2pt minus 2pt}
\setlength{\abovedisplayskip}{6pt plus 2pt minus 2pt}
\setlength{\belowdisplayskip}{6pt plus 2pt minus 2pt}
\setlength{\abovecaptionskip}{4pt}
\setlength{\belowcaptionskip}{1pt}
\makeatletter
\renewcommand\subsection{\@startsection{subsection}{2}{\z@}
  {-1.8ex plus -.5ex minus -.2ex}{.75ex plus .2ex}
  {\normalfont\large\bfseries}}
\renewcommand\section{\@startsection{section}{1}{\z@}
  {-2.1ex plus -.5ex minus -.2ex}{1.05ex plus .2ex}
  {\normalfont\Large\bfseries}}
\renewcommand\paragraph{\@startsection{paragraph}{4}{\z@}
  {0.9ex plus .3ex minus .2ex}{-1em}
  {\normalfont\normalsize\bfseries}}
\makeatother

\clubpenalty=10000\widowpenalty=10000

\begin{abstract}
We formalize Agentic Waterfall Feed, a new problem of personalized information discovery in which users actively direct recommendations and the system adapts to evolving needs inferred from both explicit requests and browsing behavior. We address this problem with the Agentic Waterfall framework, whose language-model controller maintains persistent user state and coordinates content retrieval, ranking, and state updates. To evaluate systems on this task, we contribute the Agentic Waterfall Dataset and a replay-based evaluation protocol. The dataset combines real content metadata with simulated interaction histories. The protocol compares systems under shared histories using relevance scores and blind judgments of how well recommendations satisfy current user goals and constraints. Agentic Waterfall remains competitive with strong baselines on conventional recommendation tasks across five public benchmarks. On the Agentic Waterfall Dataset, it outperforms the strongest baseline in both evaluations, with relative improvements of up to 16.1\%.
\end{abstract}

\begin{figure*}[!t]
\centering
\includegraphics[width=0.94\textwidth,trim=28bp 48bp 22bp 98bp,clip]{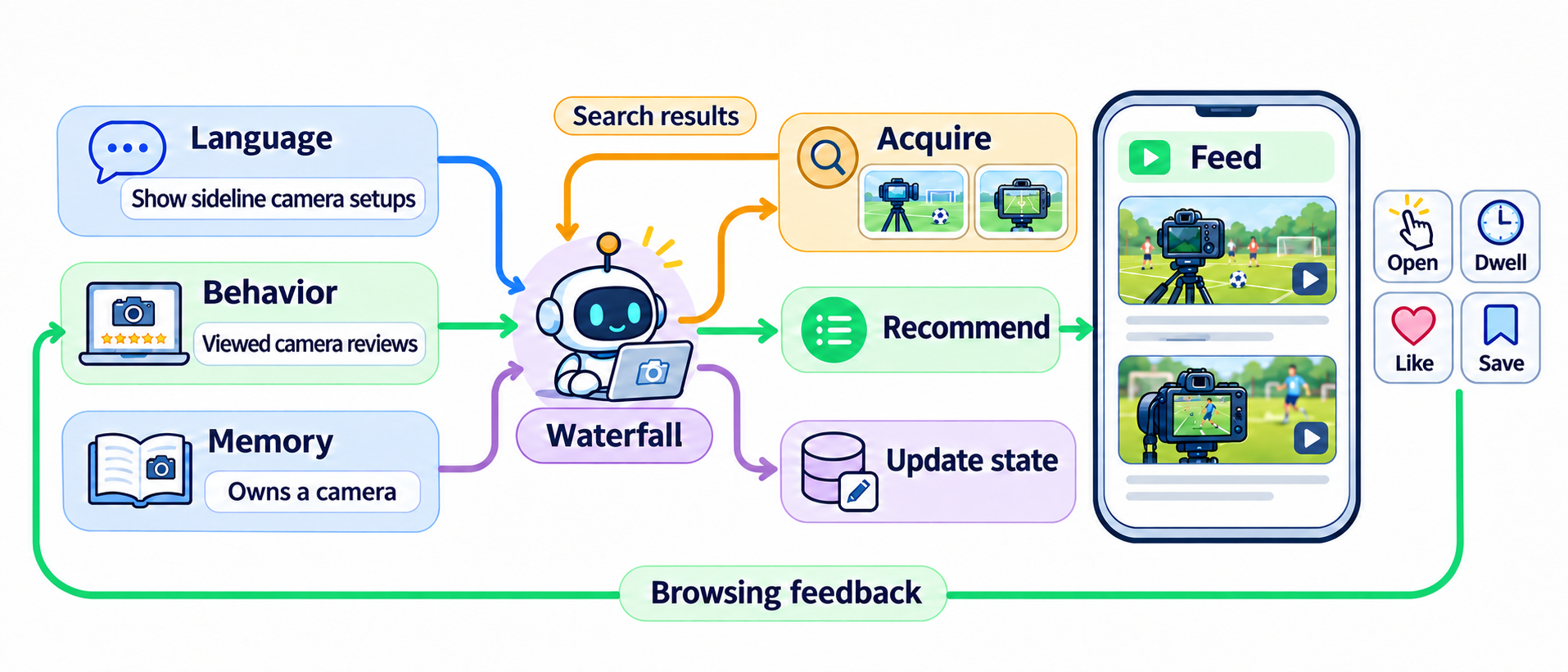}
\caption{Overview of the Agentic Waterfall framework. A language-model controller combines the current request with browsing behavior and memory to choose among content acquisition, recommendation, and state updates. Search results and user responses to the feed inform subsequent decisions.}
\label{fig:waterfall-method}
\end{figure*}

\section{Introduction}
In the attention economy, recommendation feeds shape what users see, while platforms determine the objectives governing those feeds. These objectives can diverge from users' current goals \citep{xu-etal-2025-iagent}. Users therefore need a way to redirect recommendations as their needs change. For example, a user may move from comparing cameras to seeking filming guides after making a purchase.

We formalize this requirement as \emph{Agentic Waterfall Feed}. At each interaction, a system uses the available requests and browsing history to infer the user's current goal and constraints, then decides what information to acquire and recommend.

Building on this task formulation, we develop the Agentic Waterfall framework, whose language-model controller coordinates retrieval, ranking, and state updates. It combines capabilities studied in prior work on user commands in RecBot \citep{tang2026recbot}, hierarchical memory in MARS \citep{shen2026mars}, and retrieval planning in ClawRec \citep{wu2026clawrec}. The controller links these capabilities through a persistent record of user facts, preferences, and active goals, which it updates from requests and browsing evidence to guide subsequent retrieval and recommendation.

To evaluate how well a system follows these changing needs, we introduce a replay protocol that compares systems under the same interaction histories. It combines relevance scoring and blind comparisons by a model judge, using the current goal and constraints to assess the recommendations. However, such adaptation is difficult to evaluate from conventional click or rating datasets alone.

We therefore contribute the Agentic Waterfall Dataset, which combines real content metadata with simulated interaction histories. We define two targeted subsets, \emph{Behavior Ambiguity} and \emph{Interactive Revision}, to address two recurring challenges in existing recommendation systems. Replaying these histories supports end-to-end comparison of different recommendation implementations. We compare ten systems spanning traditional and LLM-based methods, evaluate five Agentic Waterfall variants, and complement these experiments with evaluations on five public benchmarks.

\section{The Waterfall method}
\label{sec:method}

\subsection{Task formulation}
Agentic Waterfall Feed is a sequential information-discovery task. At interaction $t$, the system receives a current natural-language request $c_t$, if any, together with a history $h_{\leq t}$ of earlier requests and browsing events observed so far. It uses this evidence to infer the user's current goal $g_t$ and produce an ordered list $L_t=(i_{t,1},\ldots,i_{t,k})$ of content that supports that goal.

To construct $L_t$, the controller selects actions $A_t$ as
\begin{equation}
o_t = (c_t,h_{\leq t},m_t,e_t), \qquad A_t = \pi(o_t),
\end{equation}
where $m_t$ is persistent user memory, $e_t$ describes available content and tools, and $\pi$ is the language-model controller.

Conventional sequential recommenders such as SASRec \citep{kang2018sasrec} predict the next item from interaction histories and do not directly accept free-form requests $c_t$ as an input. Responding to $c_t$ requires understanding its meaning in the context of the user's history, which motivates LLM-based approaches.

\subsection{Action selection and execution}
\begingroup
\begin{itemize}[topsep=4pt,itemsep=3pt,parsep=0pt,partopsep=0pt]
\item \textsc{Acquire} searches a permitted source or requests details about an item. Returned content and metadata enter the candidate pool.
\item \textsc{Recommend} orders items from the available pool and replaces or appends to the pending feed queue.
\item \textsc{UpdateState} writes an evidence-linked change to the user records, such as marking a task complete or revising a constraint.
\end{itemize}

\begin{figure*}[!t]
\centering
\includegraphics[width=0.87\textwidth,trim=6bp 10bp 6bp 10bp,clip]{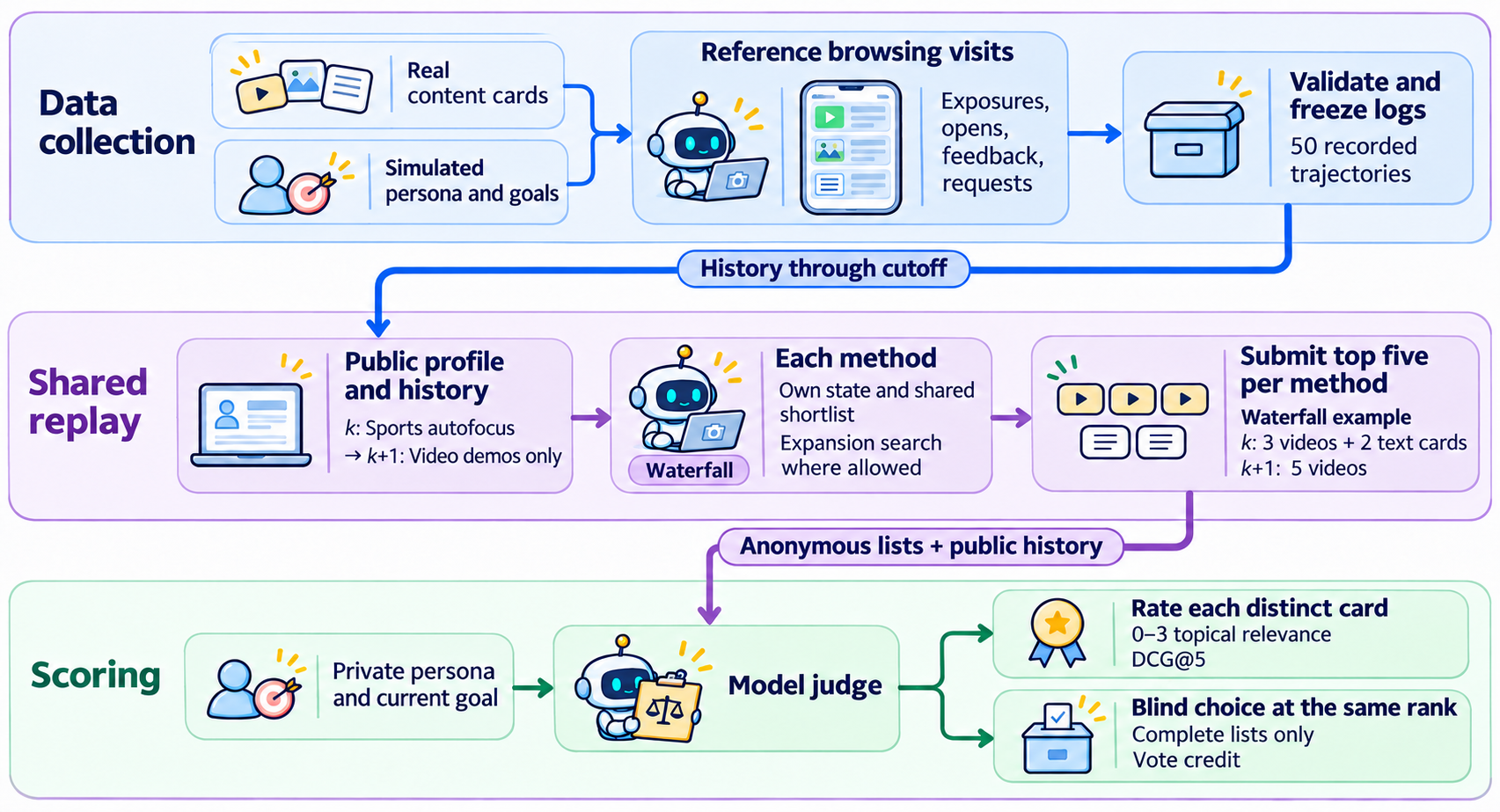}
\caption{Shared-history replay on the Agentic Waterfall Dataset. Replaying a simulated user's recorded trajectory provides the same interaction history to every system at each checkpoint. In the example, each system recommends five items at $k$ and again at $k+1$ after the user narrows a request about sports autofocus. The judge uses the private persona and current goal to rate individual cards and compare anonymous cards at the same rank.}
\label{fig:shared-history}
\end{figure*}

Agentic Waterfall brings user understanding, memory updates, retrieval, and ranking under a single language-model controller (Figure~\ref{fig:waterfall-method}). The controller interprets new requests and browsing behavior in the context of memory $m_t$, which stores user facts, preferences, and active goals. Based on this understanding, it decides which action or combination of actions to take. For example, the controller may select \textsc{UpdateState} to revise memory when new requests or browsing behavior change its understanding of the user's needs.

The results of acquisition in turn inform the controller's next decision. The controller checks how well the returned content supports the current goal. It may search further, revise the recommendations, or combine the two. Once recommendations are shown, subsequent browsing and new requests provide fresh evidence about the user's needs (see the example below). We therefore introduce a new protocol in Section~\ref{sec:protocol} to evaluate systems end to end on the Agentic Waterfall Feed task as user needs evolve across interactions.

\par\vspace{6pt}\noindent\textbf{Example.}\quad
After buying a camera, a user may ask for advice on photographing a football match. The request changes the recommendation target from choosing equipment to learning how to use it, even though the browsing history still consists of camera comparisons.

\par\endgroup

\section{Datasets}
\label{sec:datasets}

\subsection{Public benchmarks}
We use five public benchmarks to evaluate existing recommendation tasks before turning to the evolving needs studied in Agentic Waterfall Feed.

\begin{itemize}[leftmargin=1.15em,labelsep=.45em,topsep=2pt,itemsep=1pt,parsep=0pt,partopsep=0pt]
\item \textbf{AgentRecBench} \citep{shang2025agentrecbench} contains recommendation tasks from Amazon, Goodreads, and Yelp, including classical, evolving-interest, and cold-start settings. We evaluate 300 tasks with 20 candidates each.
\item \textbf{KuaiRand} \citep{gao2022kuairand} records video viewing behavior, including randomly exposed videos. We rank 101 fixed candidates in each of 617 tasks, with one held-out long-view item as the target.
\item \textbf{KuaiRec} \citep{gao2022kuairec} provides dense user-video interactions. We evaluate 1,129 users on a shared frozen catalog, requiring systems to find useful candidates as well as rank them.
\item \textbf{ClawRec-SimBench} \citep{wu2026clawrec} evaluates next-step recommendations from cross-platform behavior histories against a user's held-out next need. We evaluate all 210 events, with systems retrieving content before ranking it.
\item \textbf{MIND} \citep{wu2020mind} is a news recommendation dataset with logged candidate impressions and click histories. We evaluate systems on the MIND-small development set.
\end{itemize}

These benchmarks evaluate ranking and next-step recommendation from observed histories. Agentic Waterfall Feed also requires tracking how recommendations respond when similar browsing histories reveal different needs or an explicit request changes the current goal. This response is hard to trace in the public benchmarks because they lack controlled trajectories of such goal shifts. We therefore construct a dataset with these comparisons built into its user journeys.

\subsection{Agentic Waterfall Dataset}
The Agentic Waterfall Dataset combines real content metadata with simulated interaction histories. The catalog contains 60,674 distinct cards from 25 sources across multiple content formats, including short clips, long-form videos, articles, and illustrated posts. To model how people browse in practice, we generate each trajectory with a language-model agent guided by a private user persona and a short-term goal. Across feed visits, the trajectories record which items users open, like, or save and how long they spend on feed cards and linked content. Its 50 trajectories and 300 evaluation checkpoints cover general browsing and two targeted subsets, \emph{Behavior Ambiguity} and \emph{Interactive Revision}.

\emph{Behavior Ambiguity} contains ten pairs of users with distinct personas. The two users in each pair initially see the same cards and take the same actions, so their early browsing history fits either persona. As they pursue different goals on later visits, their choices of what to open or save begin to diverge. These users make no explicit requests or comments, so their different needs must be inferred from browsing behavior. \emph{Interactive Revision} contains twenty users who each revise an initial request, switching tasks or narrowing or relaxing requirements. We place checkpoints on either side of each revision, holding browsing history and the supplied profile fixed across each pair.

The remaining trajectories record general browsing without scripted revisions or paired behavior. Replaying each user's full trajectory gives all systems the same interaction history at each checkpoint and lets us compare their performance over time. The following protocol specifies how each system receives this evidence and how its recommendations are evaluated.

\section{Evaluation protocol}
\label{sec:protocol}

We use Hit@$K$, Precision@$K$, nDCG@$K$, AUC, and MRR to compare rankings with the relevance labels for each public-benchmark task. On ClawRec-SimBench, a large language model assigns these labels by scoring recommendations without seeing which system submitted them. We likewise use a language model as a blind judge in our end-to-end evaluation of Agentic Waterfall Feed, rating item relevance and comparing recommendations at each checkpoint to measure how well each system's feed serves the user's goals.

\subsection{Shared-history replay}
Evaluation checkpoints let us compare what different systems would show the same user at the same moment (Figure~\ref{fig:shared-history}). At each checkpoint, we pause replay and collect the top five items from each system given the same public profile and interaction history. Replay then continues with the next recorded interactions, and systems with memory carry their records forward.

Using this replay protocol, we compare Agentic Waterfall with local adaptations of ClawRec, RecAgent, and iAgent \citep{wu2026clawrec,shang2025agentrecbench,xu-etal-2025-iagent}, as well as BM25+BGE, which combines lexical and dense scores and reranks with BGE \citep{robertson2009bm25,chen-etal-2024-m3}. Because these methods differ in whether they can actively search for content, all of them start from a shared candidate pool, while methods that support acquisition can search a larger content pool within our fixed catalog. This setup lets us fairly compare the end-to-end performance of each implementation without introducing extra modules that could bias the results. All language-model methods and the judge use DeepSeek-V4.1-Flash-1017. To evaluate the returned lists at each checkpoint, the judge receives the interaction history along with the private persona and short-term goal that guided the simulated user's browsing.

\subsection{Relevance scoring}
Based on this context and each card's metadata, the judge assigns a topical relevance score from 0 to 3 to each distinct recommended card. We aggregate the top-five ratings for each system using discounted cumulative gain with linear gains \citep{jarvelin2002cumulated},
\begin{equation}
  \mathrm{DCG}@5 = \sum_{j=1}^{5}\frac{r_j}{\log_2(j+1)},
  \label{eq:dcg}
\end{equation}
where $r_j$ is the topical rating at rank $j$. Scores are averaged over each method's fully rated top-five lists.

\begin{table*}[!t]
\centering
\scriptsize
\setlength{\tabcolsep}{2.4pt}
\begin{tabular}{@{}lrrrrrrrrrr@{}}
\toprule
& AgentRecBench & \multicolumn{2}{c}{KuaiRand} & \multicolumn{2}{c}{KuaiRec} & SimBench & \multicolumn{4}{c}{MIND}\\
\cmidrule(lr){2-2}\cmidrule(lr){3-4}\cmidrule(lr){5-6}\cmidrule(lr){7-7}\cmidrule(l){8-11}
Method & Hit@5 & Hit@10 & nDCG@10 & P@10 & nDCG@10 & nDCG@20 & AUC & MRR & nDCG@5 & nDCG@10\\
\midrule
\rowcolor{waterfallrow}
Waterfall & 63.67 & 27.39 & \underline{0.1597} & \textbf{9.30} & \textbf{0.1092} & \textbf{0.7790} & \underline{0.6592} & \textbf{0.3298} & \textbf{0.3630} & \textbf{0.4214}\\
Generic agent & \textbf{64.33} & 28.04 & 0.1461 & \underline{7.73} & \underline{0.0778} & \underline{0.7084} & 0.6500 & 0.3195 & 0.3517 & 0.4110\\
ClawRec & \underline{64.00} & \underline{30.47} & 0.1595 & 4.98 & 0.0503 & 0.3450 & 0.6391 & 0.3112 & 0.3427 & 0.4015\\
RecAgent & \underline{64.00} & -- & -- & -- & -- & -- & 0.6182 & 0.3098 & 0.3385 & 0.3948\\
iAgent & 54.00 & 20.42 & 0.1056 & -- & -- & -- & 0.5985 & 0.2789 & 0.3005 & 0.3621\\
MARS & 57.33 & \textbf{32.58} & \textbf{0.1799} & -- & -- & -- & 0.6370 & 0.3070 & 0.3370 & 0.3967\\
RecBot (semantic) & -- & 11.18 & 0.0542 & -- & -- & -- & -- & -- & -- & --\\
BM25+BGE & 39.67 & -- & -- & 5.46 & 0.0537 & 0.6226 & -- & -- & -- & --\\
TF-IDF & 37.33 & 10.53 & 0.0522 & 4.84 & 0.0500 & 0.5040 & 0.5646 & 0.2681 & 0.2845 & 0.3438\\
NRMS & -- & -- & -- & -- & -- & -- & \textbf{0.6636} & \underline{0.3202} & \underline{0.3522} & \underline{0.4156}\\
\bottomrule
\end{tabular}
\caption{Public benchmark results. Hit and precision are percentages. A dash indicates an unreported result. \textbf{Bold} and \underline{underline} mark the two highest distinct values in each column.}
\label{tab:public}
\end{table*}

\begin{table*}[!t]
\centering
\small
\setlength{\tabcolsep}{5.5pt}
\begin{tabular}{@{}lrrrrrrr@{}}
\toprule
& \multicolumn{3}{c}{Overall (300)} & \multicolumn{2}{c}{Behavior ambiguity (102)} & \multicolumn{2}{c}{Interactive revision (152)}\\
\cmidrule(lr){2-4}\cmidrule(lr){5-6}\cmidrule(l){7-8}
Method & Mean relevance & DCG@5 & Votes & DCG@5 & Votes & DCG@5 & Votes\\
\midrule
\rowcolor{waterfallrow}
Waterfall & \textbf{2.64} & \textbf{7.86} & \textbf{419.58} & \textbf{7.20} & \textbf{134.50} & \textbf{8.40} & \textbf{229.75}\\
ClawRec & \underline{2.55} & \underline{7.53} & \underline{361.33} & 6.92 & \underline{109.08} & \underline{8.04} & \underline{189.75}\\
iAgent & 2.51 & 7.42 & 232.92 & \underline{7.03} & 95.83 & 7.71 & 101.92\\
RecAgent & 2.47 & 7.33 & 201.92 & 6.53 & 75.33 & 7.98 & 94.25\\
BM25+BGE & 2.16 & 6.42 & 144.25 & 5.80 & 47.25 & 6.91 & 72.33\\
\bottomrule
\end{tabular}
\caption{Results on the Agentic Waterfall Dataset. Headers show scheduled checkpoints. Blind comparisons allocate 1,360 vote credits across the five methods. \textbf{Bold} and \underline{underline} mark the two highest values in each column.}
\label{tab:pilot}
\end{table*}

\subsection{Blind comparison}
Relevance scoring evaluates each item separately. We complement it with rank-by-rank blind comparisons that test which system's recommendation the user would most want to open given their current goal and constraints.\footnote{Format compliance is checked against the recorded card metadata.} At checkpoints where all five systems return complete lists, the judge compares the shuffled cards at each rank. Each chosen card earns one vote, shared among the methods that submitted it at that rank. Ties, responses of neither, and indeterminate answers receive no vote credit.

\section{Results}
\label{sec:results}

We first compare systems on the public recommendation tasks (Table~\ref{tab:public}), including a generic agent baseline that prompts the same language model to search and rank candidates with free-form memory, and then evaluate their end-to-end performance on the Agentic Waterfall Dataset under the shared-history replay protocol (Table~\ref{tab:pilot}).

\subsection{Public benchmarks}
\label{sec:public}

\begin{itemize}[leftmargin=1.15em,labelsep=.45em,topsep=2pt,itemsep=1pt,parsep=0pt,partopsep=0pt]
\item \textbf{AgentRecBench} Waterfall hits 191 targets in the top five, two fewer than the generic agent and one fewer than ClawRec and RecAgent, matching the top fixed-candidate rankers.
\item \textbf{KuaiRand} MARS has the highest Hit@10 and nDCG@10. Waterfall ranks second in nDCG@10 by placing hit items at higher ranks.
\item \textbf{KuaiRec} Waterfall leads both metrics: top-ten precision reaches 9.3\%, compared with 7.7\% for the generic agent, and nDCG@10 reaches 0.1092 versus 0.0778.
\item \textbf{ClawRec-SimBench} Waterfall scores 0.78 nDCG@20, ahead of the generic agent at 0.71.
\item \textbf{MIND} Waterfall has the highest reported MRR (0.3298), nDCG@5 (0.3630), and nDCG@10 (0.4214), while NRMS \citep{wu2019nrms} leads AUC (0.6636). The generic agent ranks second among the agent methods on all four metrics.
\end{itemize}

\subsection{Agentic Waterfall Dataset}
\label{sec:pilot-results}

Across the public benchmarks, Agentic Waterfall matches the strongest baselines on fixed-candidate ranking and achieves clear gains on tasks that require retrieving new candidates. On the Agentic Waterfall Dataset (Table~\ref{tab:pilot}), Agentic Waterfall ranks first overall in both evaluations, reaching 7.86 in topical DCG@5 (+4.4\% over ClawRec and +22.4\% over BM25+BGE) and 419.58 blind vote credits (30.9\% of the total, +16.1\% over ClawRec's 361.33).

\begin{table*}[!t]
\centering
\footnotesize
\setlength{\tabcolsep}{3.2pt}
\begin{tabular}{@{}lrrrrrrrrr@{}}
\toprule
& \multicolumn{3}{c}{Overall (300)}
& \multicolumn{3}{c}{Behavior ambiguity (102)}
& \multicolumn{3}{c}{Interactive revision (152)} \\
\cmidrule(lr){2-4}
\cmidrule(lr){5-7}
\cmidrule(l){8-10}
Method
& Mean & Median & SD
& Mean & Median & SD
& Mean & Median & SD \\
\midrule
\rowcolor{waterfallrow}
Full Waterfall
& \textbf{7.856} & 8.415 & 1.542
& \textbf{7.198} & 7.984 & 2.056
& \textbf{8.396} & 8.845 & 0.794 \\
\texttt{no\_expression}
& 6.786 & 7.528 & 2.171
& 6.259 & 7.169 & 2.342
& 6.959 & 7.678 & 2.127 \\
\texttt{no\_behavior}
& 6.467 & 7.528 & 2.447
& 4.414 & 3.722 & 2.269
& 8.345 & 8.459 & 0.826 \\
\texttt{fixed\_workflow}
& 7.313 & 8.214 & 2.050
& 5.952 & 6.828 & 2.609
& 8.231 & 8.845 & 0.976 \\
\texttt{no\_acquire}
& 7.511 & 8.214 & 1.753
& 6.539 & 7.349 & 2.215
& 8.313 & 8.459 & 0.777 \\
\texttt{frozen\_state}
& 7.434 & 8.072 & 1.820
& 6.250 & 6.997 & 2.267
& 8.325 & 8.459 & 0.789 \\
\bottomrule
\end{tabular}
\caption{Ablation results on the Agentic Waterfall Dataset. We report mean, median, and standard deviation of DCG@5 on fully rated lists. The headers give scheduled checkpoints. Bold marks the highest mean in each group.}
\label{tab:ablation}
\end{table*}

\paragraph{Behavior Ambiguity.}
When users make no explicit requests and diverge only through browsing actions, Agentic Waterfall achieves the highest topical DCG@5 (7.20) and 134.50 blind votes, ahead of both ClawRec (6.92 DCG@5, 109.08 votes) and iAgent (7.03 DCG@5, 95.83 votes). Tracking browsing evidence in persistent memory helps the controller separate users whose trajectories begin similarly but later pursue different goals.

\paragraph{Interactive Revision.}
The gain is especially clear when users explicitly revise their requests: Agentic Waterfall reaches 8.40 in DCG@5 and 229.75 blind votes, outperforming ClawRec (8.04 DCG@5, 189.75 votes) by 21.1\% in blind votes. Immediately after a request revision, Agentic Waterfall averages 8.52 in DCG@5 compared with 8.10 for ClawRec, and satisfies explicit format constraints on all five recommended cards at 39 of 40 constrained checkpoints (versus 34 for ClawRec).

\section{Ablation study}
\label{sec:ablation}
We replayed five Agentic Waterfall variants across the checkpoints on the Agentic Waterfall Dataset (Table~\ref{tab:ablation}). Two omit either user expression or browsing. The others fix action order, disable content acquisition, or block state writes. The variants start from the same public profile and history prefix, then rebuild state and candidates with the allowed inputs at each checkpoint. Full uses the archived lists from the shared-history replay, which carried state across checkpoints.

In Table~\ref{tab:ablation}, Full scored 7.198 in behavior ambiguity and the variant without browsing scored 4.414. In interactive revision, Full scored 8.396 and the variant without user expression scored 6.959. The latter variant also scored below Full in behavior ambiguity (6.259 vs.\ 7.198), where users gave no requests or comments. The difference in state carryover could contribute to these gaps.

\begin{figure}[!t]
\centering
\includegraphics[width=\columnwidth,trim=12bp 46bp 12bp 132bp,clip]{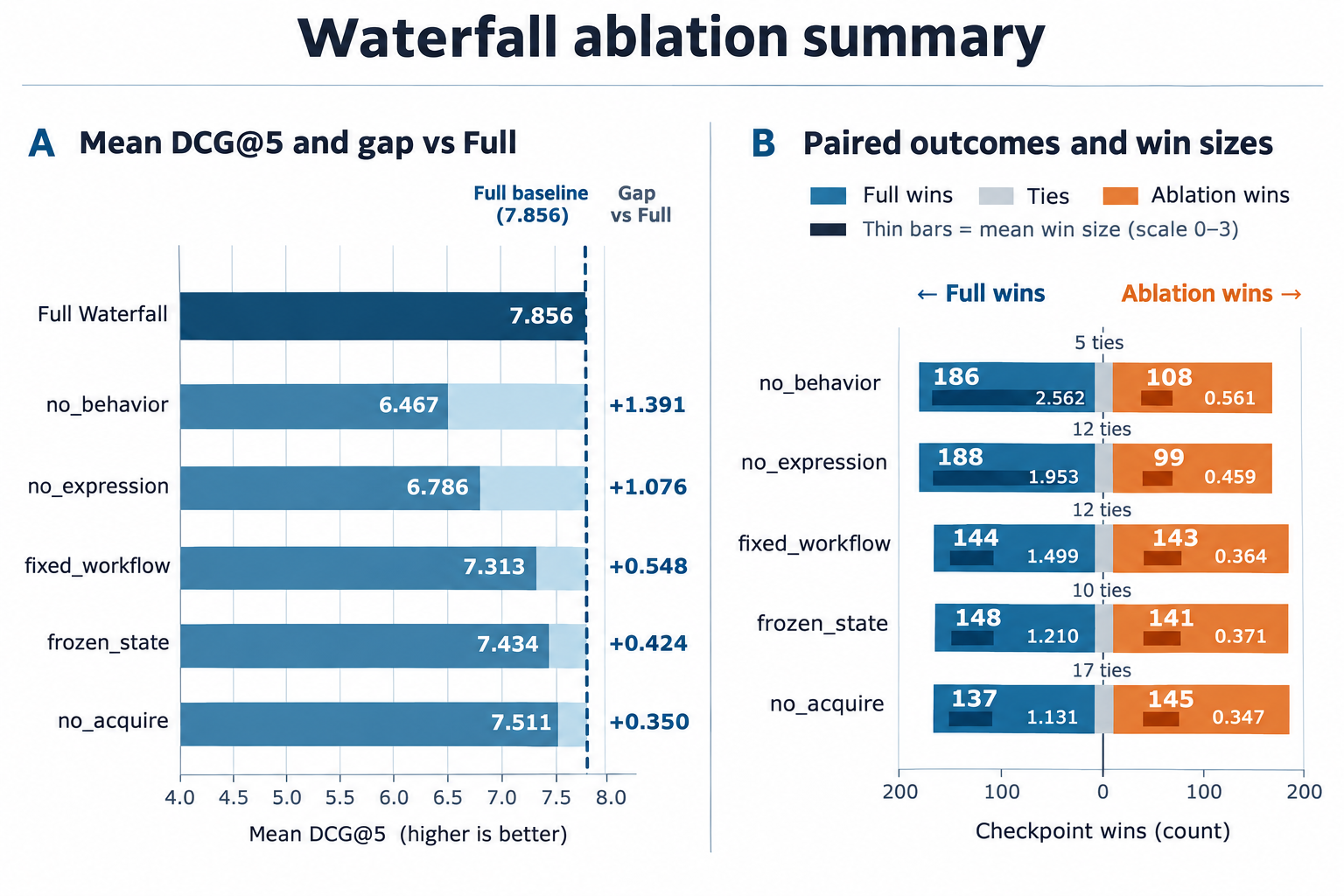}
\caption{Comparison of each ablation with Full Waterfall. Panel (a) gives mean DCG@5 by configuration and the paired mean difference (Full minus variant). Panel (b) gives wins, ties, and the mean winning margin on each side. The bars use each configuration's scorable lists. Paired differences and win counts use 299 jointly scorable checkpoints per comparison.}
\label{fig:ablation-summary}
\end{figure}

Full had roughly as many wins as the fixed workflow, no-acquisition, and frozen-state variants (Figure~\ref{fig:ablation-summary}). With content acquisition disabled, the variant won 145 checkpoints and Full won 137, with 17 ties. Full's larger winning margins yielded a positive mean gap of 0.350.

\section{Conclusion}
We formalize Agentic Waterfall Feed and present Agentic Waterfall, an information-discovery framework whose language-model controller coordinates content acquisition, recommendation, and state updates as user needs evolve. We also contribute the Agentic Waterfall Dataset and a shared-history replay protocol to evaluate systems end to end through relevance scoring and blind comparisons. Across five public benchmarks and our dataset, Agentic Waterfall remains competitive on conventional ranking tasks, leads on retrieval-based tasks, and achieves the best overall performance under both behavioral ambiguity and explicit request revisions.

\paragraph{Code and data availability.}
The \href{https://github.com/XiaoLuoLYG/Agentic-Waterfall}{\textcolor{urlblue}{\underline{code}}} and \href{https://github.com/XiaoLuoLYG/agentic-waterfall-dataset}{\textcolor{urlblue}{\underline{data}}} are available on GitHub, including trajectories, catalog manifests, checkpoint inputs, run records, and blinded evaluation labels.

\paragraph{Ethics.}
The experiments reported in this study used simulated users, model-generated interactions, and model judgments. Although our demo system supports user data collection, no collected user data is included in this study or publicly released.

\begingroup
\interlinepenalty=10000
\bibliographystyle{plainnat}
\bibliography{references}
\endgroup
\par\endgroup
\end{document}